\pdfoutput=1

\documentclass[11pt]{article}

\usepackage[final]{acl}

\usepackage{times}
\usepackage{latexsym}

\usepackage[T1]{fontenc}

\usepackage[utf8]{inputenc}

\usepackage{microtype}

\usepackage{inconsolata}

\usepackage{graphicx}
\usepackage{booktabs}
\usepackage{multirow}
\usepackage{graphicx}
\usepackage{makecell}
\usepackage{xcolor}
\usepackage{amsmath}
\title{MTPChat: A Multimodal Time-Aware Persona Dataset \\ for Conversational Agents}

\author{
 \bfseries
 Wanqi Yang$^{1 \ast}$
 \
 Yanda Li$^{1}$\thanks{Equal contributions}
 \
 Meng Fang$^{2}$
 \
 Ling Chen$^1$
 \\
 \normalsize 
 $ ^1$ University of Technology Sydney
 \
 $ ^2 $ University of Liverpool
 \\
 {\normalsize \tt   wanqi.yang-1@student.uts.edu.au, 
 \normalsize \tt Yanda.Li@student.uts.edu.au}\\
 \normalsize \tt  Meng.Fang@liverpool.ac.uk,
 \normalsize \tt ling.chen@uts.edu.au
 }

\begin{document}
\maketitle



\begin{abstract}
Understanding temporal dynamics is critical for conversational agents, enabling effective content analysis and informed decision-making. However, time-aware datasets, particularly for persona-grounded conversations, are still limited, which narrows their scope and diminishes their complexity. To address this gap, we introduce MTPChat, a multimodal, time-aware persona dialogue dataset that integrates linguistic, visual, and temporal elements within dialogue and persona memory. Leveraging MTPChat, we propose two time-sensitive tasks: Temporal Next Response Prediction (TNRP) and Temporal Grounding Memory Prediction (TGMP), both designed to assess a model’s ability to understand implicit temporal cues and dynamic interactions. Additionally, we present an innovative framework featuring an adaptive temporal module to effectively integrate multimodal streams and capture temporal dependencies. Experimental results validate the challenges posed by MTPChat and demonstrate the effectiveness of our framework in multimodal time-sensitive scenarios.
\end{abstract}

\section{Introduction}
Temporal awareness has garnered significant attention in AI research, particularly following Min et al.’s work \cite{min2020ambigqa}, which highlighted the inherent temporal dynamics in question-answering systems. Understanding time-sensitive information is crucial across various domains, including financial decision-making, event prediction, multimedia content analysis, and conversational AI. To explore temporal reasoning in large language models (LLMs), multiple time-sensitive datasets have been developed. TimeQA \cite{chen2021dataset} and SituatedQA \cite{zhang2021situatedqa} provide temporally grounded questions with free-text contexts extracted from WikiData \cite{vrandevcic2014wikidata}. Similarly, TEMPLAMA \cite{dhingra2022time} builds on temporal knowledge bases, while StreamingQA \cite{liska2022streamingqa} compiles time-sensitive question-answering (QA) data from English news articles spanning 2007 to 2020.

\begin{figure}[t]
\centering
\includegraphics[width=\columnwidth]{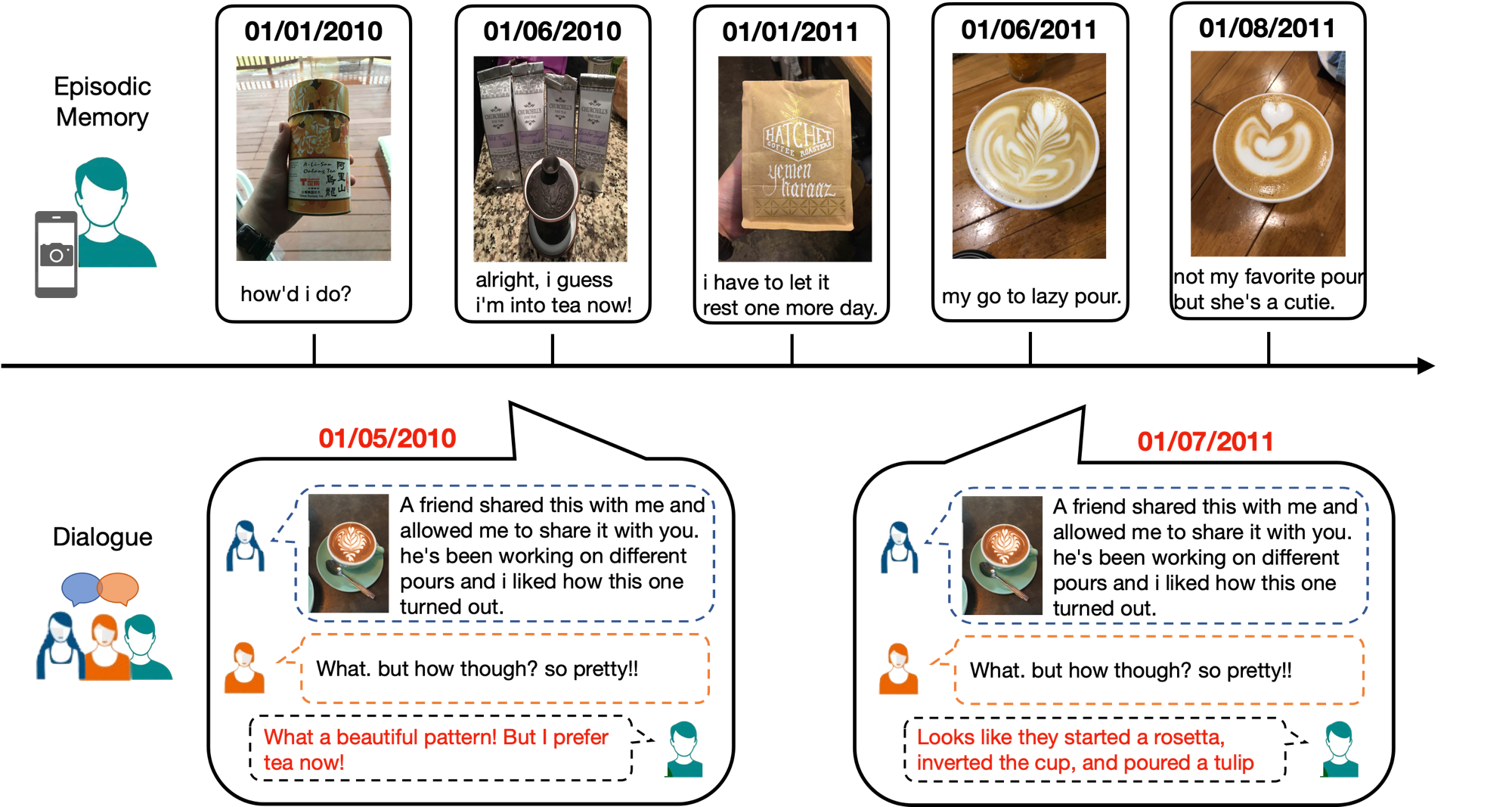}
\caption{An example of a multimodal, time-sensitive, persona-grounded scenario, showcasing how the user's dialogue responses evolve over time based on the temporal dynamics of dialogue and episodic memories.
}
\vspace{-3mm}
\label{intro}
\end{figure}

\begin{table*}[ht]
\centering
\resizebox{0.95\linewidth}{!}
{
\begin{tabular}{l|ccccc}
\toprule
\textbf{Dataset} & \textbf{Knowledge Corpus} & \textbf{\# Samples} & \textbf{Time-Sensitive} & \textbf{Task} & \textbf{has Images}\\
\midrule
TempLama~\cite{dhingra2022time} & CustomNews & 50.0k & YES & Question Answering & NO\\
TimeQA~\cite{chen2021dataset} & Wikipedia & 41.2k & YES & Question Answering & NO\\
StreamingQA~\cite{liska2022streamingqa} & WMT07-20 & 138.0k & YES & Question Answering & NO\\
TempReason-L2L3~\cite{tan2023towards} & Wikipedia & 49.0k & YES & Question Answering & NO\\

\midrule
PhotoChat~\cite{zang2021photochat} & OpenImage V4& 12.3k & NO & Dialogue & YES \\
MMDialog~\cite{feng2022mmdialog} & SocialMedia& 1.1M & NO & Dialogue & YES\\
\midrule
\textbf{MTPChat} &Reddit & 28.7k & YES & Dialogue & YES \\
\bottomrule
\end{tabular}
}
\vspace{-2mm}
\caption{\label{comparison}
Related datasets overview, including free-text time-sensitive datasets and multimodal dialogue datasets.}
\vspace{-4mm}
\end{table*}


However, these datasets primarily focus on text-based QA tasks, limiting their applicability to real-world conversational AI. They lack the multimodal components (e.g., images) that are essential for capturing rich temporal contexts and do not account for persona-grounded dialogues, where responses evolve based on a user's dynamic memory and past interactions. Although TimeIT \cite{ren2023timechat} introduces time-sensitive multimodal tasks for long-video understanding, it has several limitations: (1) its focus on QA tasks restricts broader conversational applications, (2) the use of explicit temporal markers in videos reduces the challenge of reasoning over implicit temporal cues, and (3) its rigid response format  (e.g., ``<timestamp\_start> to <timestamp\_end> seconds: <event\_description>'') simplifies the task, minimizing complex temporal reasoning. Similarly, MPChat~\cite{ahn2023mpchat} provides persona-grounded dialogues with multimodal memory, but it lacks an explicit temporal dimension.


To overcome these limitations, we introduce MTPChat, a multimodal, time-aware persona dialogue dataset built upon MPChat~\cite{ahn2023mpchat}, a comprehensive multimodal persona-grounded dialogue dataset. Rather than relying on explicit timestamps, MTPChat leverages the natural progression of dialogues and memories to simulate real-world temporal shifts in human cognition. Figure~\ref{intro} illustrates an example of a multimodal time-sensitive scenario. Our dataset integrates linguistic, visual, and temporal elements, making it the first of its kind to model persona-driven temporal changes in dialogues and memory. Unlike existing time-sensitive datasets, MTPChat incorporates dialogue, persona memory, and visual elements, enhancing its realism and complexity. In addition, we propose two novel tasks—Temporal Next Response Prediction (TNRP) and Temporal Grounding Memory Prediction (TGMP)—that challenge models to infer implicit temporal cues and track evolving responses over time.



Beyond dataset creation, we introduce an adaptive temporal module designed to enhance the temporal reasoning capabilities of multimodal models. This framework dynamically integrates linguistic, visual, and temporal streams, allowing for more effective reasoning over time-sensitive interactions. Specifically, the module dynamically merges features based on their temporal relevance, improving coherence in multimodal integration.



To evaluate MTPChat, we conducted experiments using SBERT \cite{reimers2019sentence} and CLIP \cite{radford2021learning}. Our results demonstrate that MTPChat introduces new challenges in multimodal, time-sensitive scenarios, requiring models to track temporal changes effectively.
Our adaptive temporal module outperforms other feature integration methods, significantly enhancing a model’s ability to reason over multimodal time-aware dialogue.

The main contributions of this work are as follows:
 \begin{itemize}
 
    \item We introduce the first multimodal, time-aware persona dialogue dataset, which contains numerous instances where both dialogue responses and grounding memories evolve significantly over time.
 
    \item We define Temporal Next Response Prediction (TNRP) and Temporal Grounding Memory Prediction (TGMP) to advance research in time-aware conversational AI.

    \item We propose a framework with an adaptive temporal module that enhances a model’s ability to integrate multimodal streams while maintaining temporal awareness.
  
    \item Experimental results validate the novel challenges posed by MTPChat and demonstrate that our framework outperforms existing methods in multimodal temporal reasoning.

\end{itemize}

\section{Comparison with Existing Datasets}
We start with a brief comparison of existing datasets, emphasizing multi-modal and time-aware strategies (see Table~\ref{comparison} for an overview).

\paragraph{Time-Sensitive QA Datasets}

Time-Sensitive Question Answering (TSQA) involves interpreting and responding to questions that are dependent on specific time points or intervals. We analyse a set of TSQA datasets~\cite{dhingra2022time,chen2021dataset,liska2022streamingqa,tan2023towards}, as shown in the upper part of Table~\ref{comparison}. 
Currently, TSQA datasets typically use free-text form or knowledge graphs (KGs) and are structured as QA tasks. However, our work introduces the first multimodal time-aware dataset based on conversation. Similar to TSQA, we modify the time of dialogues, which affects the responses and the related grounding memory, thereby testing the model's ability to understand time.

\paragraph{MultiModal Dialogue Datasets}

Multimodal dialogue datasets generally comprise one or more images and multi-turn textual dialogues. As depicted in the lower half of Table~\ref{comparison}, we analyse two representative datasets~\cite{zang2021photochat,feng2022mmdialog}. These datasets are designed for models to interpret images and utterances within a dialogue framework and generate coherent responses. Our MTPChat dataset, although drawing on the conversational structure and task, distinctively emphasizes the annotation and manipulation of time information. MTPChat allows the model to acknowledge the influence of temporal dynamics on dialogue interaction and memory processes, demonstrating temporal awareness. 

\paragraph{Time-Sensitive Video-Centric Dataset}

TimeIT~\cite{ren2023timechat} is a novel dataset focused on video-based instructions, encompassing a collection of long-video datasets annotated with timestamps. It requires models to describe video content across specified time intervals. The description follows a structured format, such as ``<timestamp\_start> to <timestamp\_end> seconds: <event\_description>''. 
Ingeniously, our dataset integrates time of dialogues and memories, making model awareness of the time order of dialogue and memory significant influence on dialogue responses and memory recall. In contrast to TimeIT's tasks that directly answer timestamp and associated content, MTPChat offers a more complex challenge with implicit time factor, pushing the boundaries of temporal understanding in multimodal dialogue models.

\section{MTPChat Dataset}

Our dataset is built on the basis of MPChat~\cite{ahn2023mpchat}, a comprehensive multimodal persona-grounded dialogue dataset that includes both linguistic and visual components derived from episodic-memory-based personas. MPChat gathered from the social media platform Reddit, consists of memory image-sentence pairs and dialogue instances grounded on the speakers' multimodal memories. 

A significant challenge is the ingenious integration of time information and multimodal dialogue, aiming to establish a multimodal time-aware dataset. Based on MPChat dataset, we develop a novel methodology that involves three primary steps: 1) Time annotations, 2) Constructing time-aware conversations, and 3) Memory annotations. These efforts achieve the creation of a pioneering multimodal time-aware dialogue dataset. MTPChat breaks away from the limitations of current time-sensitive datasets confined to QA tasks, free-text formats and relying on explicit time information. We believe that our work fosters the development of more diverse time-sensitive datasets and advancing research toward achieving human-level temporal understanding in models.

\subsection{Time Annotations}
\label{section:2.1}

We converted the UTC strings in MPChat dataset into date format ``yyyy/mm/dd'' and incorporated this feature into both the dialogue and memory components. The dialogue in our dataset is structured as a triplet consisting of (dialogue context, dialogue image, dialogue time), while each memory of the speaker is similarly organized as a triplet (memory description context, memory image, memory time).

\subsection{Time-Aware Conversations}
\label{section:2.2}

In real-world scenarios, conversations can vary significantly based on the time they occur, even with similar contexts. For instance, as a high school student asked, ``What is machine learning?'', you might respond with no knowledge on the subject. However, after three years of studying machine learning at university, your response to the same conversation would be more detailed, potentially including discussions about deep learning and related topics.

Inspired by how the temporal order of conversation and memories influences human responses, we constructed conversational data with temporal orders:
\begin{itemize}
\vspace{-2mm}
\item Later Stage Conversations: We used the original memories and conversations from the MPChat dataset, adding time annotations as described in Section~\ref{section:2.1}. For instance, if you are a university student with three years of study in machine learning and are asked, ``What is machine learning?'', your response might include topics like deep learning.
\vspace{-1mm}
    \item Early Stage Conversations: To simulate conversations from earlier times, we assumed there was no prior memory of the discussion topic. We used the context of the original conversations but removed the original responses. We then add new, earlier time annotations and responses. The newly created responses differ from the original ones and contain minimal information about the discussion topic due to the lack of relevant memory. For example, if you are a high school student asked, ``What is machine learning?'', you might respond with little to no knowledge on the subject.

    Specifically, we utilized GPT-4 \cite{achiam2023gpt} to process a combination of inputs: the dialogue context, dialogue image, newly modified dialogue time, and speaker memories predating this new dialogue time. GPT-4 generated responses under the following guidelines: 1) responses could not exceed 40 words; 2) if the provided memories' topics significantly differed from the conversation, the response should indicate the speaker’s lack of familiarity with the conversations topic; 3) if the provided memories and conversation topics were only slightly different, the response should reflect the speaker's intention to engage with and explore the conversation topic.
\end{itemize}

\subsection{Memory Annotations}

\vspace{-1mm}
To gain a more precise understanding of the model's capabilities in temporal awareness, we align conversations with memory. For the memory component, we add time annotations as outlined in Section~\ref{section:2.1}. Since the memories of the speakers are sourced from real users on Reddit, we avoid creating fabricated memories to preserve data authenticity. Additionally, we incorporate a ``No Memory'' category into the speaker's memory set. Structured similarly to existing memory triplets (memory description context, memory image, memory time), the ``No Memory'' category is assigned as the description context, indicating that there is no memory to align with the response. \footnote{We also correlate ``No Memory'' with a plain white image as the memory image.} This memory category is used to align early-stage conversations. We then synchronize the memory time with the conversation's time information.

\subsection{Dataset Statistics}

\begin{figure}[t]
\centering
\includegraphics[width=0.85\columnwidth]{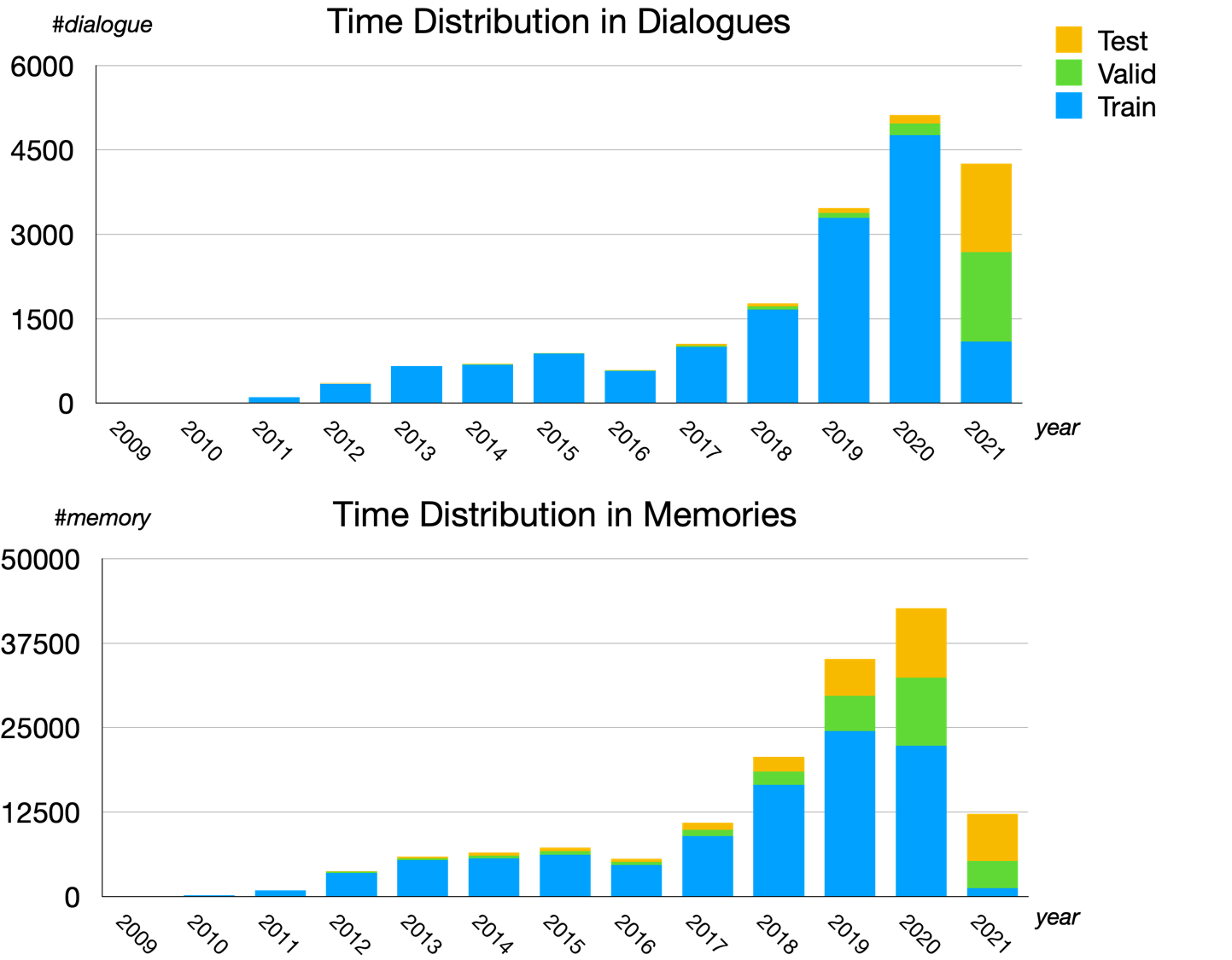}
\vspace{-2mm}
\caption{Distribution of times across conversations and memories in training, validation, and test set.
}
\vspace{-2mm}
\label{data}
\end{figure}

MTPChat comprises 18,973 conversations and 25,877 users. We divided MTPChat into training, validation, and test sets with 15,056, 1,994, and 1,923 conversations respectively. We analyzed the proportion of later stage conversations and early stage conversations, finding a ratio of 3:1. As well as later stage conversations with grounding memories (some later stage conversations lack grounding memory) and early stage conversations with ``No Memory'', resulting in a ratio of 2:1. Furthermore, to gain deeper insight into the time information within MTPChat, we charted the distribution of times across conversations and memories in Fig~\ref{data}.

\section{Task Definition}

\begin{figure*}[h]
\centering
  \includegraphics[width=0.95\textwidth]{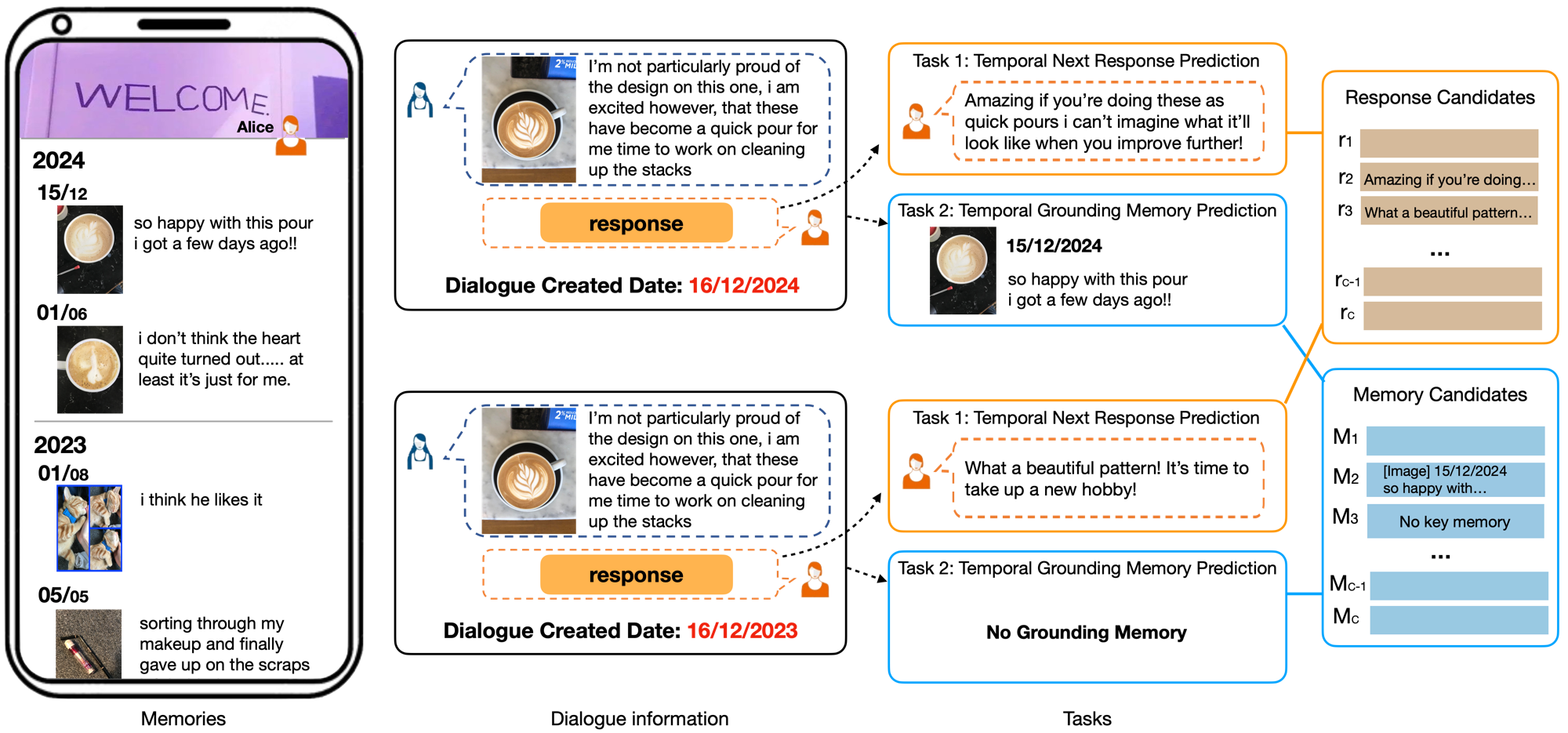}
  \vspace{-2mm}
  \caption{Overview of the Temporal Next Response Prediction (TNRP) and Temporal Grounding Memory Prediction (TGMP) tasks. The left panel displays a user’s episodic memories, represented as image-sentence-time triplets with various creation dates. The dialogue instance on the right highlights the corresponding response and task setup. }
    \label{task}
\vspace{-4mm}
 \end{figure*}


The MTPChat datasets consist of N examples $\mathcal{D}=\{({d}_n, r_n, \mathcal{M}_n)\}_{n=1}^N$, where \( \forall n \in \{1, \ldots, N\} \) and each example contains a dialogue ${d}_n$, the speaker's response ${r}_n$ to the dialogue ${d}_n$ and a memory set $\mathcal{M}_n$ from the speaker. Each dialogue $d_n=(c^{d_n}, {i}^{d_n}, t^{d_n})$ encompasses the context $c^{d_n}$ (context utterances), an associated image ${i}^{d_n}$ and the time 
marking ${t}^{d_n}$ (formatted as yyyy/mm/dd) when the dialogue occurred. The memory set for the speaker consists of m distinct memories $\mathcal{M}_n=\{M_{n_1},\ldots,M_{n_m}\}$, where each memory $M_{n_m} = (c^{M_{n_m}}, i^{M_{n_m}}, t^{M_{n_m}})$ characterized by a description context $c^{M_{n_m}}$ (context utterances), an image $i^{M_{n_m}}$ and the time marking $t^{M_{n_m}}$ (formatted as yyyy/mm/dd) when the memory occurred.

\subsection{Temporal Next Response Prediction}

As illustrated in the Fig~\ref{task}, Temporal Next Response Prediction (TNRP) is a retrieval task aimed at predicting the next response $\tilde{r}$ from a set $R_c$ containing $C$ response candidates based on the dialogue $d=(c^{d}, {i}^{d}, t^{d})$ and the speaker's memories $\mathcal{M}=\{M_{1}=(c^{M_{1}}, i^{M_{1}}, t^{M_{1}}),\ldots,M_{m}\}$. The response candidate set $R_c$ comprises one ground truth and $C-1$ distractor responses. It is essential to emphasize that, 1) Identical dialogue content and speaker memories can lead to vastly different responses depending on the time of the dialogue. 2) To intensify the task's complexity and underline the temporal factor's significance, our response candidate set includes responses from later-stage dialogue and early-stage dialogue. The remainder of the response candidates are randomly selected from other dialogues. 


\subsection{Temporal Grounding Memory Prediction}

Temporal Grounding Memory Prediction (TGMP) task is also a retrieval task that requires predicting the most likely memory element from a set $M_c$ containing $C$ memory candidates based on a given dialogue $d=(c^{d}, {i}^{d}, t^{d})$ and a remainder memory set (except grounding memory) before producing a response. The memory candidate set $M_c$ comprises one grounding memory, one ``No Memory'' option and $C-2$ distractor memories randomly selected from other speakers. As shown in Fig~\ref{task}, time variations within the dialogue substantially influence the choice of the grounding memory. Specifically, when the time of the dialogue is later than the time of the grounding memory, suggesting the availability of memory related to the dialogue for supporting the speaker’s response, the model is capable of predicting the grounding memory. Conversely, if the time of the dialogue is earlier than that of the grounding memory, indicating an absence of relevant dialogue memory, the model must predict a ``No Memory'' outcome.

In TGMP task, we deliberately exclude the speaker's response from the input. This decision is based on the consideration that potential responses of early-stage dialogue can vary significantly—ranging from disinterest in the dialogue topic to expressing a desire to learn. These different but reasonable responses could potentially confuse the model to predict grounding memory. The principal objective of the TGMP task is making model recognize the critical temporal order between dialogue and memory. 
By focusing on whether the model can identify the appropriate grounding memory or its absence for a given time information, we obtain a clearer measure of its temporal awareness capabilities.
\begin{figure}[t]
\centering
\includegraphics[width=0.9\columnwidth]{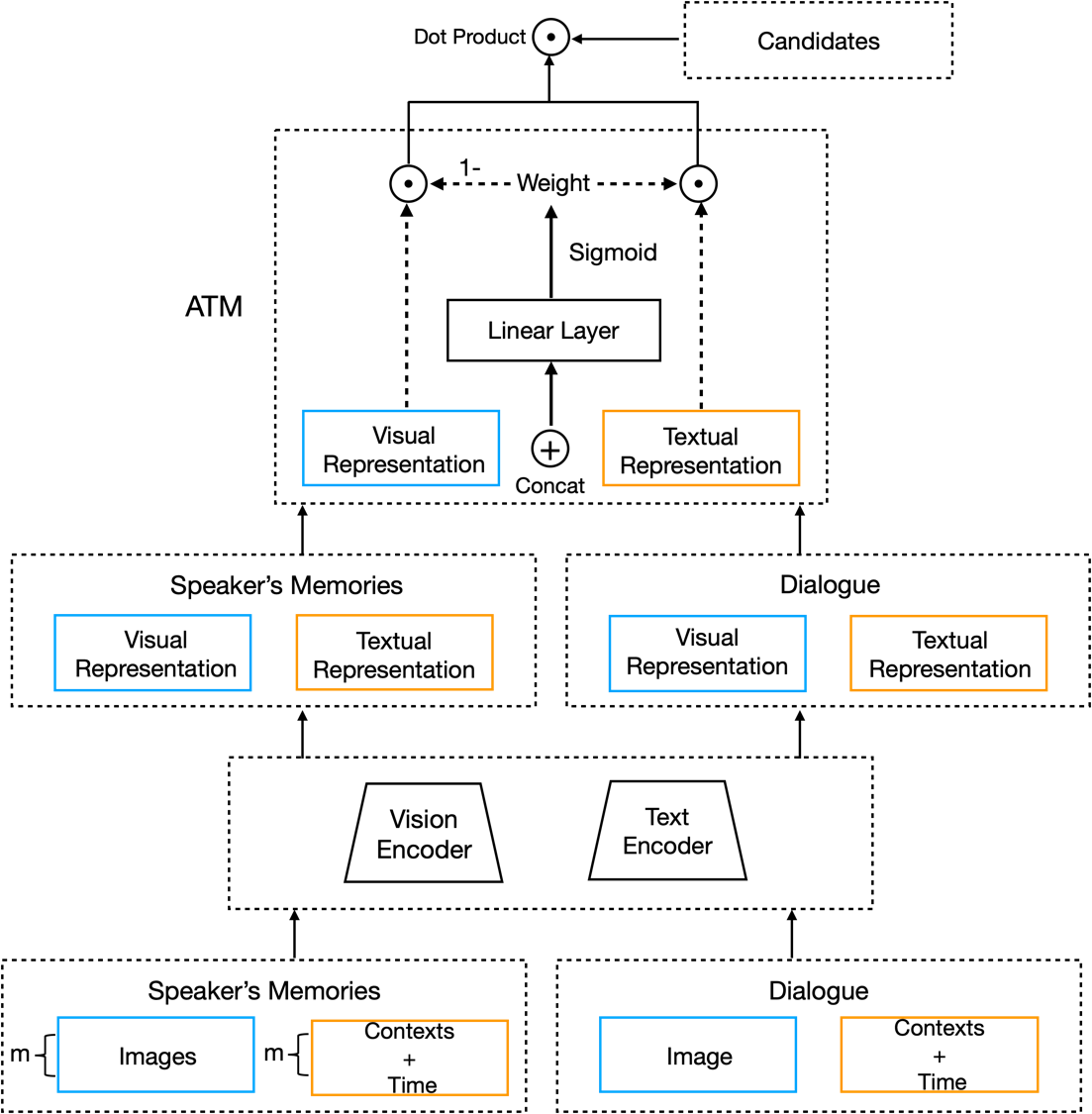}
\vspace{-2mm}
\caption{Architecture of our framework with Adaptive Temporal Module (ATM).
}
\label{model}
\vspace{-4mm}
\end{figure}
\vspace{-2mm}
\section{Framework}
In this section, we present a framework to perform above retrieval tasks based on dialogue and memory. The inputs include dialogue ${d}_n$, the speaker's response ${r}_n$ to the dialogue and a memory set $\mathcal{M}_n$. We define various encoders to process different modalities of data, fuse the extracted features, and achieve both the temporal next response prediction task and the temporal grounding memory prediction task. The architecture of our framework is shown in Fig~\ref{model}.

\textbf{Text Encoder}
In this study, we employ the text encoder to process textual components within tasks, specifically extracting representations of text and date information from dialogues, memories, and responses. For both dialogue and speaker memories, which may contain multiple entries, we first concatenate the text and date information for each entry. These concatenated strings are then further combined using a delimiter, forming unified representations. This method ensures comprehensive feature extraction by the text encoder, facilitating a more robust analysis of the textual data involved.

\textbf{Vision Encoder}
Similarly, our vision encoder to extract features from images embedded in dialogues and memories. In datasets featuring speaker memories with multiple images, each image is processed by this vision encoder. The extracted features are then aggregated via mean-pool operation to create a consolidated visual representation. This methodology ensures a coherent integration of visual data, significantly enhancing the model's capacity to process multi-image features effectively.

\textbf{Adaptive Temporal Module}
Following the extraction of textual and visual representations, it is essential to effectively integrate these features. As the inclusion of date information into textual representations can impact the correspondence between the text and vision features extracted by text encoder and vision encoder, we propose a method to dynamically balance these modalities to maintain the alignment of text and visual information within the same set of memories and dialogues. We introduce a module called the Adaptive Temporal Module (ATM), which is designed to be both simple and effective.

First, we concatenate the corresponding text and vision features and map them through a linear layer. Subsequently, a sigmoid layer is used to derive the weights for both text and vision features. These weights are then employed to merge the features based on their relevance, ensuring better alignment and integration. This approach allows for a more coherent and contextually appropriate fusion of multimodal features, enhancing the overall interpretative capability of the model.

\section{Experiments}
\subsection{Experimental Setup}
\paragraph{Baselines} We consider the following baselines:

\begin{itemize}
\vspace{-2mm}
    \item SBERT+CLIP: We adopt a Transformer~\cite{vaswani2017attention} initialized weights of SBERT~\cite{reimers2019sentence} and CLIP-ViT-B/32 vision model~\cite{radford2021learning} as text encoder and vision encoder to represent text and image respectively. SBERT enhances the original BERT model~\cite{devlin2018bert} to better handle similarity comparisons of  dialogue and memory textual information. CLIP-ViT-B/32 vision model utilizes a Vision Transformer (ViT)~\cite{dosovitskiy2020image} with 32 attention heads, which enables it to capture more visual features.
    
\vspace{-2mm}
    \item CLIP+CLIP: We utilize the CLIP-ViT-B/32 model~\cite{radford2021learning} as text encoder (CLIP-ViT-B/32 text model) and vision encoder (CLIP-ViT-B/32 vision model). CLIP-ViT-B/32 text model employs a Transformer similar to GPT~\cite{radford2018improving}, designed specifically for processing textual input, making it ideally suited for textual analysis requirements.

\end{itemize}

\paragraph{Training}

We train both baselines and our framework for 5 epochs with a batch size of 8 on a NVIDIA Tesla V100 GPU. The model is optimized using Adam~\cite{kingma2014adam} with a learning rate of $3e^{-6}$. For our framework, we incorporated the Adaptive Temporal Module (ATM) into two baselines to validate the effectiveness of framework. We set the number of speaker's memories is $m=20$ and the number of candidates is $C=100$.

\paragraph{Evaluation Metrics}
We assess the performance of the model on two tasks using Recall@1 and Mean Reciprocal Rank (MRR), which is the standard evaluation metrics on dialogue task~\cite{lee2021constructing,feng2022mmdialog, ahn2023mpchat}. Recall@1 quantifies the model's accuracy in retrieving the most relevant result as the top result for each query, effectively capturing the model's ability to return the most relevant result as the first item. MRR evaluates the average inverse ranking of the first relevant result across queries, providing insight into the model's overall retrieval quality.

\subsection{Experimental Results}

We conduct experiments of two baselines with and without our framework on time-sensitive tasks in MTPChat. Besides, we define two input settings: one limited to dialogue, and the other encompassing both dialogue and speaker's memories. The findings, as depicted in Table~\ref{results1}, reveal several insights: 1) MTPChat poses challenges in terms of the multimodal temporal awareness capabilities of models. Despite TNRP and TGMP being retrieval tasks, both baselines exhibited inadequate performance on these time-sensitive challenges, achieving Recall@1 scores not surpassing 70. 2) Our framework is model-agnostic and effective, enhancing performance over both baselines. Note that in our TNRP task, where labels contain only the response text, the ATM module—which is tailored for multimodal fusion balance—yields a less pronounced improvement. 
3) The temporal ordering of dialogue and memories plays a pivotal role in MTPChat. In previous works with multimodal persona-grounded dialogue datasets~\cite{zhong2020towards,wen2021automatically}, the persona information serves as supplementary data to improve the accuracy of predicted dialogue responses. However, in MTPChat, both persona memory and dialogue are essential components. They not only enhance the model's temporal awareness but also significantly influence performance. For instance, for CLIP+CLIP+ATM model on TGMP task, when the input lacked memory data, performance significantly dropped by 20.1\% in Recall@1 and 15.1\% in MRR.

\begin{table}[t]

\centering
\resizebox{\linewidth}{!}{
\begin{tabular}{l|c|c|c|c|c}
\toprule
\multirow{2}*{Model} & \multirow{2}*{Input Setting} &\multicolumn{2}{c|}{TNRP} & \multicolumn{2}{c}{TGMP}\\ 
\cline{3-6}
 \multirow{2}{*}{}&\multirow{2}{*}{} &R@1&MRR&R@1&MRR\\
\midrule
\multirow{2}*{SBERT+CLIP} &{$d$} & 58.26 & 69.90 & 49.17 & 63.38\\
\cline{2-6}
\cline{2-6}
\multirow{2}*{} &{$d$, $\mathcal{M}$} & 61.32 & 72.55 & 58.90 & 73.53 \\
\midrule
\multirow{2}*{SBERT+CLIP+ATM} &{$d$} & 58.70 & 70.26 & 52.04 & 65.35 \\
\cline{2-6}
\cline{2-6}
\multirow{2}*{} &{$d$, $\mathcal{M}$} &  61.55 & 72.78 & 60.22 & 74.26  \\
\midrule
\multirow{2}*{CLIP+CLIP} &{$d$} & 66.20 & 76.34 & 56.91 & 70.64\\
\cline{2-6}
\cline{2-6}
\multirow{2}*{} &{$d$, $\mathcal{M}$} & 68.75 & 78.66 & 67.25& 80.50\\
\midrule

\multirow{2}*{CLIP+CLIP+ATM} &{$d$} & 66.97 & 76.96 &  57.35 & 71.04 \\
\cline{2-6}
\cline{2-6}
\multirow{2}*{} &{$d$, $\mathcal{M}$} & \textbf{69.26} & \textbf{78.92} & \textbf{71.82} & \textbf{83.68} \\
\bottomrule
\end{tabular} 
}
\vspace{-2mm}
\caption{\label{results1}
Results of the Temporal Next Response Prediction (TNRP) and Temporal Grounding Memory Prediction (TGMP) tasks. Symbols means: dialogue $d=(c^{d}, {i}^{d}, t^{d})$ contains a context, an image and time information. A speaker's memory set $\mathcal{M}=\{M_{1},\ldots,M_{m}\}$, where each memory $M = (c^M, i^M, t^M)$ characterized by a context, an image and time information.
}
\end{table}


\begin{table}[t]

\resizebox{\linewidth}{!}{
\begin{tabular}{p{4.5cm}|p{3cm}<{\centering}|p{3cm}<{\centering}}
\toprule
\multirow{2}*{Method} & \multicolumn{2}{c}{Temporal Grounding Memory Prediction}\\ 
\cline{2-3}
 \multirow{2}{*}{} &R@1&MRR\\
\midrule
Attention Fusion& 63.65 & 76.72\\
\midrule
Linear Fusion& 66.41 & 79.59 \\
\midrule
Mean-Pool Fusion& 67.25 & 80.50 \\
\midrule
ATM (ours)& \textbf{71.82} & \textbf{83.68} \\
\bottomrule
\end{tabular} 
}
\vspace{-2mm}
\caption{\label{results2}
Comparison of Adaptive Temporal Module (ATM) with other methods of feature integration on Temporal Grounding Memory Prediction task.
}
\vspace{-3mm}
\end{table}

In addition, to evaluate the performance of the Adaptive Temporal Module within our proposed system, we conducted a comparative analysis against other feature fusion methods:
\begin{itemize}
 \vspace{-2mm}
    \item Attention Fusion: This method adeptly combines textual and temporal data with image features, employing an attention-based module to learn weights. This enhances the model's sensitivity to contextually significant features.
 \vspace{-2mm}   
    \item Linear Fusion: Incorporates two linear layers optimized during training, enabling the model to directly learn the weights that most effectively combine textual and visual information.
 \vspace{-6mm} 
    \item Mean-Pool Fusion: This approach computes the mean of the combined features, aggregating them from different modalities by simple averaging.

\end{itemize}

These methods were assessed using the CLIP+CLIP model on the Temporal Grounding Memory Prediction (TGMP) task. The findings in Table~\ref{results2} indicate that the Adaptive Temporal Module surpassed other techniques, achieving improvements of 12.8\%, 8.1\%, and 6.4\% in Recall@1, respectively. These results substantiate the superior capability of our framework to effectively enhance multimodal integration with temporal awareness.

\subsection{Ablation Study}

\paragraph{Zero-Shot Setting}

\begin{table}[h]

\centering
\resizebox{\linewidth}{!}{
\begin{tabular}{l|c|c|c|c|c}
\toprule
\multirow{2}*{Model} & \multirow{2}*{Input Setting} &\multicolumn{2}{c|}{TNRP} & \multicolumn{2}{c}{TGMP}\\ 
\cline{3-6}
 \multirow{2}{*}{}&\multirow{2}{*}{} &R@1&MRR&R@1&MRR\\
\midrule
\multirow{2}*{CLIP+CLIP} &{$d$, $\mathcal{M}$(zero-shot)} & 39.49 & 52.07 & 54.59 & 61.27 \\
\cline{2-6}
\cline{2-6}
\multirow{2}*{} &{$d$, $\mathcal{M}$} & 68.75 & 78.66 & 67.25 & 80.50  \\

\bottomrule
\end{tabular} 
}
 \vspace{-2mm} 
\caption{\label{results3}
Ablation study of baseline CLIP+CLIP with zero-shot setting.
}
\vspace{-2mm} 
\end{table}

We explore the performance of the CLIP+CLIP model with a zero-shot setting on time-sensitive tasks. As shown in Table~\ref{results3}, the model demonstrates poor performance on MTPChat time-sensitive tasks, showing the challenges inherent in MTPChat and highlighting the urgent need for research into multimodal temporal awareness.

\paragraph{The Importance of Temporal Awareness}

\begin{table}[h]

\centering
\resizebox{\linewidth}{!}{
\begin{tabular}{l|c|c|c}
\toprule
\multirow{2}*{Model} & \multirow{2}*{Input Setting}  & \multicolumn{2}{c}{TGMP}\\ 
\cline{3-4}
 \multirow{2}{*}{}&\multirow{2}{*}{} &R@1&MRR\\
\midrule
\multirow{2}*{CLIP+CLIP} &{$d$, $\mathcal{M}$(without time)} & 60.99 & 65.09  \\
\cline{2-4}
\cline{2-4}
\multirow{2}*{} &{$d$, $\mathcal{M}$} & 68.75 & 78.66   \\

\bottomrule
\end{tabular} 
}
\vspace{-2mm} 
\caption{\label{results4}
Ablation study of baseline CLIP+CLIP without time information.
}
\vspace{-2mm} 
\end{table}

This study highlights the critical role of temporal awareness in models. Utilizing the CLIP+CLIP model, we trained on datasets both with and without temporal data of dialogue and memories. These models were then evaluated on the Temporal Grounding Memory Prediction (TGMP) task. Our findings (see Table~\ref{results4}) reveal a marked difference in performance: models without temporal awareness demonstrated substantial difficulties in time-sensitive tasks. Conversely, models incorporating temporal awareness significantly excelled, achieving a 12.7\% increase in recall@1 and a 20.8\% improvement in MRR.


\section{Related Work}

\paragraph{Time-Sensitive Datasets}
In recent years, time-sensitive datasets have predominantly been designed for question answering tasks and primarily consisting of textual data~\cite{zhang2021situatedqa,chen2021dataset,tan2023towards,liska2022streamingqa,wei2023menatqa,yang2024continual}. Among these, the SituatedQA dataset~\cite{zhang2021situatedqa} represents a significant contribution by emphasizing open-domain, time-sensitive question answering. It reannotates questions from the Natural Questions (NQ)~\cite{kwiatkowski2019natural} and Wikidata~\cite{vrandevcic2014wikidata} to capture contextual dependencies and temporal variations in answers. Similarly, TimeQA~\cite{chen2021dataset} comprises 20,000 questions, including a challenging variant that requires models to infer answers from implicit temporal cues in text passages. Additionally, the TempReason dataset~\cite{tan2023towards} offers a comprehensive framework for evaluating various facets of temporal understanding. In these Open Book Question Answering (OBQA) settings, models leverage external textual resources to deduce correct answers~\cite{izacard2020leveraging,zaheer2020big,wei2021finetuned,ouyang2022training,yang2024enhancing}.

Time-sensitive datasets have also been developed for Closed Book Question Answering (CBQA), where models must generate answers relying solely on the information contained in the question~\cite{fevry2020entities,roberts2020much,dhingra2022time}. Furthermore, datasets built on knowledge graphs—such as TEQUILA~\cite{jia2018tequila}, TimeQuestions~\cite{jia2021complex}, and CronQuestions~\cite{saxena2021question}—pose more complex natural language queries, requiring models to rank entities according to their temporal relevance.

\paragraph{Multimodal Dialogue Datasets}
Multimodal dialogue research has gained traction with the emergence of datasets that integrate images with multi-turn textual dialogues. Such datasets aim to jointly model visual and linguistic information to either answer questions~\cite{antol2015vqa,das2017visual,seo2017visual,kottur2019clevr,li2023stablellava} or generate coherent responses~\cite{meng2020openvidial,zheng2021mmchat,wang2021openvidial,zang2021photochat,feng2022mmdialog}. For example, Mostafazadeh et al.~\cite{mostafazadeh2017image} introduced the IGC dataset, which comprises 4,000 dialogues centered around an image accompanied by a textual description and related questions and responses. Building on this, Shuster et al.~\cite{shuster2018image} released the ImageChat dataset, a substantially larger collection that captures more diverse conversational scenarios.
Recent efforts have incorporated persona information to foster more personalized interactions. Datasets such as FoCusd~\cite{jang2022call}, MPChat~\cite{ahn2023mpchat}, DuLeMon~\cite{xu2022long}, and MSPD~\cite{kwon2023and} augment dialogues with persona details—ranging from purely textual to multimodal attributes—enabling models to extract relevant personal context and enhance the naturalness of generated responses.


\section{Conclusion}
In this work, we addressed the underexplored challenge of temporal awareness in multimodal, persona-grounded dialogues by introducing MTPChat, a multimodal, time-aware persona dialogue dataset, along with an adaptive temporal framework.
MTPChat presents new challenges by requiring conversational agents to comprehend implicit temporal dynamics in evolving dialogues and persona memories, thereby expanding the scope of temporal reasoning beyond traditional QA tasks. Additionally, our proposed adaptive temporal module has demonstrated significant improvements in model performance, underscoring its effectiveness in integrating multimodal streams and capturing dynamic temporal dependencies.
Our findings highlight the importance of temporal reasoning in conversational AI, and we anticipate that MTPChat will serve as a valuable resource for future research in multimodal, time-aware AI systems.
\section{Limitations}

Despite its comprehensive structure and innovative tasks, the MTPChat dataset and our framework present certain limitations and need attention for future development. For MTPChat dataset, while the dataset significantly enhances the challenge of temporal reasoning by incorporating implicit temporal cues, it may still not fully capture the subtleties of real-world temporal dynamics, such as those influenced by cultural, historical, or personal contexts that affect human interactions. For our framework, future research should focus on refining this framework and exploring its scalability and adaptability across different domains and temporal challenges, aiming to further our understanding of time's impact on cognitive and decision-making processes.

\section{Ethics Statement}

In the development of the MTPChat dataset, we have placed a high priority on privacy and adherence to ethical standards. We ensured that the images in the dataset do not contain identifiable features such as faces, license plates, or email addresses, and the text is free from offensive language. We urge users of the dataset to be aware of these inherent risks. Additionally, commercial use of our data is strictly limited to ensure compliance with the Reddit API Terms and to protect user privacy. The MTPChat dataset is exclusively permitted for academic research purposes.

\section{Acknowledgements}
This project is partially supported by ARC DP240101349.

\bibliography{custom}

\clearpage
\begin{center}\large\bfseries
Appendix
\end{center}

\appendix

\section{Detailed Prompt of GPT-4}
\label{sec:appendix1}

\begin{table}[h]

\centering
\resizebox{\columnwidth}{!}{
\begin{tabular}{l}
\toprule
\textbf{Prompt of GPT-4 for generating response to early-stage conversation}  \\
\midrule
Given the topic of a conversation, the context of the dialogue, and multiple memories \\of the speaker, please write a response to the conversation. \\
 \\
It is important to note:\\
1. responses could not exceed 40 words.\\
2. If the memories are almost unrelated to the conversation, the generated response \\should reflect the speaker's lack of expertise in the conversation topic. \\If appropriate, consider incorporating the current content of the speaker's memories. \\
3. If the memories are related to the conversation, the response should express \\a willingness to try or explore it in the future. \\
 \\
Conversation Topic: [topic]\\
Dialogue Context: [context]\\
Memories: [context]\\
\bottomrule
\end{tabular} }
\caption{\label{Parameters1} Detailed prompt of GPT-4 for generating response to early-stage conversation.}
\end{table}

\section{Detailed Parameters}
\label{sec:appendix2}
The parameter settings of Temporal Next Response Prediction (TNRP) and Temporal Grounding Memory Prediction (TGMP) tasks used in our paper are illustrated in Table~\ref{Parameters1}.

\begin{table}[h]

\centering
\resizebox{0.8\columnwidth}{!}{
\begin{tabular}{l|c|c}
\toprule
\textbf{Parameters} & \textbf{TNRP} & \textbf{TGMP} \\
\midrule
per\_gpu\_train\_batch\_size & 8 & 8 \\ 
per\_gpu\_eval\_batch\_size& 1 & 4  \\
num\_train\_epoch& 5 & 5  \\
max\_num\_candidates& 100 & 100   \\
max\_num\_image& 20 & 20  \\
image\_size &224 & 224 \\
learning\_rate& $3e^{-6}$ & $3e^{-6}$\\
weight\_decay &0.05 &0.05    \\
\bottomrule
\end{tabular} }
\caption{\label{Parameters1} Detailed Parameters of Temporal Next Response Prediction (TNRP) and Temporal Grounding Memory Prediction (TGMP) tasks.}
\end{table}

\end{document}